\documentclass[10pt,journal,compsoc]{IEEEtran}
\usepackage{cite}
\usepackage{amsmath,amssymb,amsfonts}
\usepackage{algorithmic}
\usepackage{graphicx}
\usepackage{textcomp}
\usepackage{comment}
\usepackage{threeparttable}
\usepackage{hyperref}
\usepackage{threeparttable}
\usepackage{multirow}
\usepackage{booktabs}
\usepackage{upgreek}
\def\BibTeX{{\rm B\kern-.05em{\sc i\kern-.025em b}\kern-.08em
    T\kern-.1667em\lower.7ex\hbox{E}\kern-.125emX}}

\begin{document}

\title{
Skeleton-Based Intake Gesture Detection With Spatial-Temporal Graph Convolutional Networks
}

\author{Chunzhuo Wang$^{1}$, Zhewen Xue$^{1}$, T. Sunil Kumar$^{2}$, Guido Camps$^{3}$, Hans Hallez$^{4}$, and Bart Vanrumste$^{1}$
\thanks{*This work is funded in part by the Horizon Europe Research and Innovation Program under Grant GA.No. 101083388, in part by the Flanders AI Research Program, and in part by the Leuven.AI Institute.}
\thanks{$^{1}$Chunzhuo Wang, Zhewen Xue, and Bart Vanrumste are with the e-Media Research Lab, and also with the ESAT-STADIUS Division, KU Leuven, 3000 Leuven, Belgium 
        {\tt\small chunzhuo.wang@kuleuven.be; bart.vanrumste@kuleuven.be}}%
\thanks{$^{2}$T. Sunil Kumar is with University of Gävle, 801 76 Gävle, Sweden
        {\tt\small sunilkumar.telagam.setti@hig.se}}%
\thanks{$^{3}$Guido Camps is with the Division of Human Nutrition and Health, Department of Agrotechnology and Food Sciences, Wageningen University and Research, 6700EA Wageningen, and also with the OnePlanet Research Center, 6708WE Wageningen, The Netherlands 
        {\tt\small guido.camps@wur.nl}}%
\thanks{$^{4}$Hans Hallez is with the M-Group, DistriNet, Department of Computer Science, KU Leuven, 8200 Sint-Michiels, Belgium 
        {\tt\small hans.hallez@kuleuven.be}}%
}

\maketitle
\thispagestyle{empty}
\pagestyle{empty}

\begin{abstract}
Overweight and obesity have emerged as widespread societal challenges, frequently linked to unhealthy eating patterns. A promising approach to enhance dietary monitoring in everyday life involves automated detection of food intake gestures. This study introduces a skeleton based approach using a model that combines a dilated spatial-temporal graph convolutional network (ST-GCN) with a bidirectional long-short-term memory (BiLSTM) framework, as called ST-GCN-BiLSTM, to detect intake gestures. The skeleton-based method provides key benefits, including environmental robustness, reduced data dependency, and enhanced privacy preservation. Two datasets were employed for model validation. The OREBA dataset, which consists of laboratory-recorded videos, achieved segmental F1-scores of 86.18\% and 74.84\% for identifying eating and drinking gestures. Additionally, a self-collected dataset using smartphone recordings in more adaptable experimental conditions was evaluated with the model trained on OREBA, yielding F1-scores of 85.40\% and 67.80\% for detecting eating and drinking gestures. The results not only confirm the feasibility of utilizing skeleton data for intake gesture detection but also highlight the robustness of the proposed approach in cross-dataset validation.
\newline

\indent \textit{Clinical relevance}— This skeleton-based intake gesture detection system can serve as an automated annotation tool for nutritional experts, facilitating the analysis of participants' eating behaviors.
\end{abstract}

\section{Introduction}
Overweight and obesity have been considered as the global health challenges. Unhealthy dietary habit is one of the causes for these health issues. Traditional dietary assessment tools, including 24 hours recall (24HR) and food diary, require manual inputs from participants, which are subjective and prone to error, therefore not suitable for long-term monitoring in daily life. Automated food intake monitoring approach that combines sensors and pattern recognition algorithms is a promising solution to tackle this limitation. 

Different sensors have been explored for this purpose including inertial measurement units (IMUs) \cite{b5,10684446,b4,10584254}, strain gauge sensors \cite{Sazonov2012}, infrared proximity sensors \cite{Bedri2015}, acoustic sensors \cite{Amft2010}, piezoelectric sensors \cite{Kalantarian2014}, photoplethysmography (PPG) sensors \cite{b16}, electromyography (EMG) sensors \cite{b17}, and radar sensors \cite{b9}. Most of these sensors are integrated in wearable devices enabling wearable based intake activity detection. Another important form is ambient-based method using video recording from cameras \cite{8853283,9187935}. Rouast and Adam \cite{8853283} proposed deep learning based approach to detect intake gestures using video data on their collected OREBA dataset. Qiu et al. \cite{9187935} used wearable camera to detect bites and recognize food types. Tang and Hoover explored the video based approach in cafeteria settings \cite{9956550}. These studies choose to use deep learning models to process RGB data directly, which may draw potential privacy concerns.

Alternatively, another type of video-based approach is to first extract human body skeleton keypoints from raw video data, then detect intake gestures by processing keypoints. This approach offers privacy protection since the participants' visual details are absent once the skeleton is extracted. Additionally, it is computationally efficient and requires less memory compared to techniques that rely on RGB video data. It is also more robust to variations in lighting, background, clothing, and other factors that may alter the visual appearance of individuals in the footage. Furthermore, using the human skeleton provides a straightforward and intuitive representation of gestures, simplifying the interpretation and understanding of classification outcomes. Tufano et al. \cite{10.3389/fnut.2024.1343868} proposed a rule-based approach to detect bites based on 3D facial keypoints. The study of \cite{7020549} extract skeleton information from Microsoft Kinect v1 RGB-D camera, and Dynamic Time Warping (DTW) algorithm is used to process the skeleton data for drinking gesture detection. The EatSense study\cite{RAZA2023104762} combines RGB and skeleton data for intake gesture classification. However it shoud be noted that these skeleton based methods mainly focus on activity classification on manually trimmed video with the assumption that each trimmed video clip contains a single activity. The more challenging scenario, skeleton-based intake gesture detection with untrimmed video, has yet to be explored. 

This study \footnote{This work is based on the master's thesis \cite{xue2023} of Zhewen Xue and Chunzhuo Wang completed at KU Leuven.} introduces a deep learning framework that leverages skeleton-based data to efficiently identify food intake gestures. The approach involves preprocessing eating video footage into skeleton data, which includes X and Y coordinate positions along with their detection confidence levels. A deep learning architecture, combining a dilated spatial-temporal graph convolutional network (ST-GCN) and bidirectional long-short-term memory (BiLSTM) model, is designed to process skeleton data as input and generate sequential outputs. The algorithm predicts classification probabilities at each time point, resulting in time-series sequences. The key contributions of this research are outlined as follows:
\begin{itemize}
    \item We proposed a deep learning pipeline to verify the feasibility of using skeleton-based
information for continuous fine-grained food intake gesture detection.
    \item We further explored the combination of different skeletal body parts to investigate the optimal solution for intake gesture detection.
    \item We evaluated the effectiveness of the skeleton-based method in both controlled environments and diverse home settings through cross-dataset validation.
\end{itemize}

\section{Method}
\subsection{Datasets}
\begin{figure}[t]
  \centering
  \includegraphics[scale=0.35]{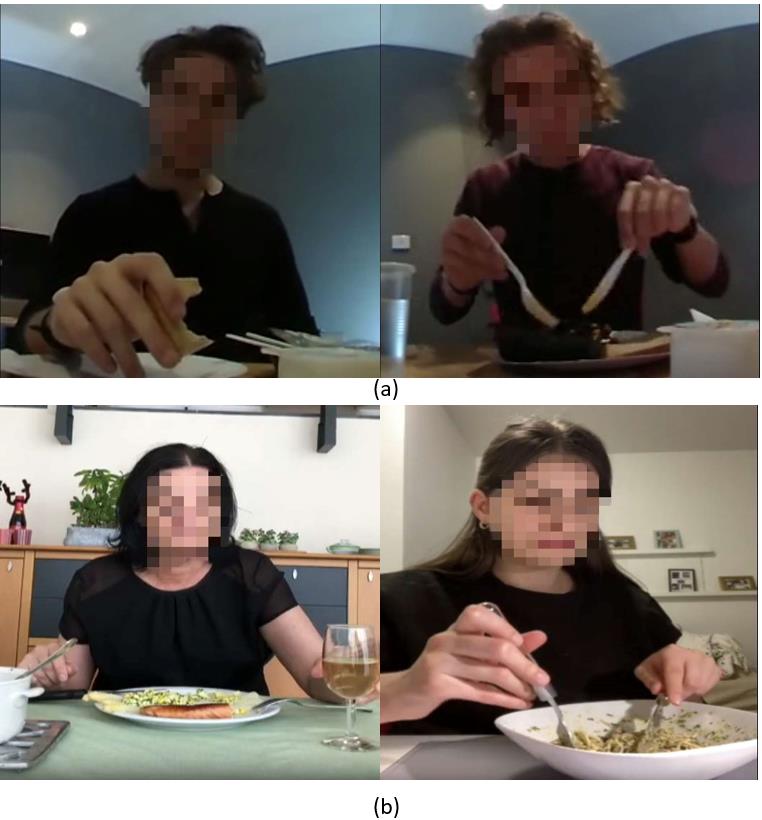}
  \caption{Data collection scenes from (a) OREBA dataset (b) Smartphone footage dataset.} 
  \label{smartphone}
\end{figure}

Two datasets are used in this study: OREBA dataset \cite{b6} and a self-collected smartphone footage dataset recorded in home settings \cite{10034636}.
\begin{itemize}
    \item \textbf{OREBA Dataset:} The OREBA dataset \cite{b6} captures group dining scenarios, where up to four individuals were filmed sharing a meal at a communal table in each session. Participants were served multiple courses, including yogurt, lasagna (available in vegetarian or beef variants), and bread. The dataset includes eating videos from 100 participants, with ages ranging from 18 to 54 years. A total of 4,496 eating and 406 drinking gestures were recorded. Recordings were made using a centrally placed spherical camera (360fly-4K5) positioned at the shared table, ensuring unobtrusive data collection. The videos were captured at 24 frames per second (FPS) with a frame resolution of 140$\times$140 pixels. This study follows the same training/validation/testing split (61/20/19) as outlined in \cite{b6}.
    \item \textbf{Smartphone Footage Dataset:} The self-collected dataset \cite{10034636} consists of smartphone-recorded videos of 20 meal sessions from 14 participants (6 males and 8 females), aged between 16 and 75 years. The recordings include 837 eating and 68 drinking gestures. Since the participants filmed themselves, variations in camera angles and distances introduced natural variability, making the dataset valuable for testing model robustness. All videos were standardized to 10 FPS. The size of each frame is reduced to 140$\times$140 pixels. This dataset serves as a test set for the model trained on OREBA training data. Dataset collection scenes are shown in Fig. \ref{smartphone}.
\end{itemize}

\subsection{Skeleton Extraction}
This paper primarily focuses on predicting eating behaviors using skeletal data. To achieve this, a publicly accessible framework was employed for skeleton extraction. Specifically, the MMPose framework \cite{mmpose2020} from OpenMMLab was utilized. MMPose is an open-source project developed collaboratively by academic institutions and industry experts. It integrates various cutting-edge algorithms for skeleton extraction and has been rigorously trained and fine-tuned on numerous datasets, showcasing superior performance in skeleton recognition tasks. For this study, we utilized a pre-trained model combining Faster R-CNN for detecting human bounding boxes and HRNet for extracting human keypoints. This model was pre-trained on the COCO-WholeBody dataset \cite{jin2020whole}, which includes annotations for 133 keypoints per individual (17 for the body, 68 for the face, 6 for the feet, and 42 for the hands). While most skeleton-based action recognition techniques rely on full-body graphs derived from natural skeletal connections \cite{wang-2022}, detecting intake gestures does not require complete body information. Moreover, full-body skeleton data may be unavailable in certain scenarios, such as when obstructions like tables block lower parts of the body. A specialized skeleton structure was developed to emphasize the connection between the hands and mouth, achieved by incorporating links between the arms, fingertips, mouth, and face (head). Since analyzing eating behavior does not necessitate all 133 keypoints, we focused on selectively extracting 23 keypoints related to specific body parts essential for this task:
\begin{itemize}
    \item \textbf{Face:} Five keypoints were considered: nose (index 1), left and right eyes (indices 2 and 3), and left and right ears (indices 4 and 5).
    \item \textbf{Arms:} Four keypoints were included: left and right shoulders (indices 6 and 7), and left and right elbows (indices 8 and 9).
    \item \textbf{Mouth:} Four keypoints were selected: rightmost (index 72) and leftmost points of the mouth (index 78), the middle of the upper lip (index 86), and the middle of the lower lip (index 90).
    \item \textbf{Hands:} Ten keypoints were identified from the thumb and index fingers. For the left hand: indices 92, 94, 96, 101, and 104. For the right hand: indices 113, 115, 117, 122, and 125.
 \end{itemize}

\begin{figure}[t]
  \centering
  \includegraphics[scale=0.30]{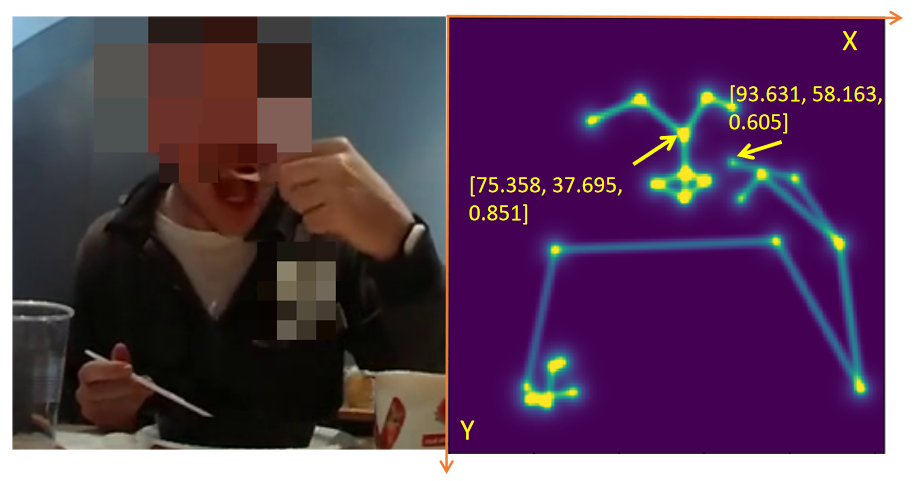}
  \caption{The original video frame alongside the reconstructed skeleton heatmap, created using preprocessed keypoint data, is presented. The heatmap includes labels for the x, y, and confidence scores corresponding to two specific keypoints.}
  \label{preprocess}
\end{figure}

These keypoints include horizontal (X) and vertical (Y) coordinates along with confidence scores. Fig. \ref{preprocess} provides an example of an original video frame alongside its corresponding skeleton representation. In the OREBA dataset, most videos feature a single participant. If multiple individuals are detected within a frame, only the skeleton keypoints of the individual with the highest confidence score are retained. In instances where certain keypoints are missing or their confidence scores fall below the predefined threshold, the corresponding x, y, and confidence values are set to 0.

\begin{figure*}[t]
  \centering
  \includegraphics[scale=0.38]{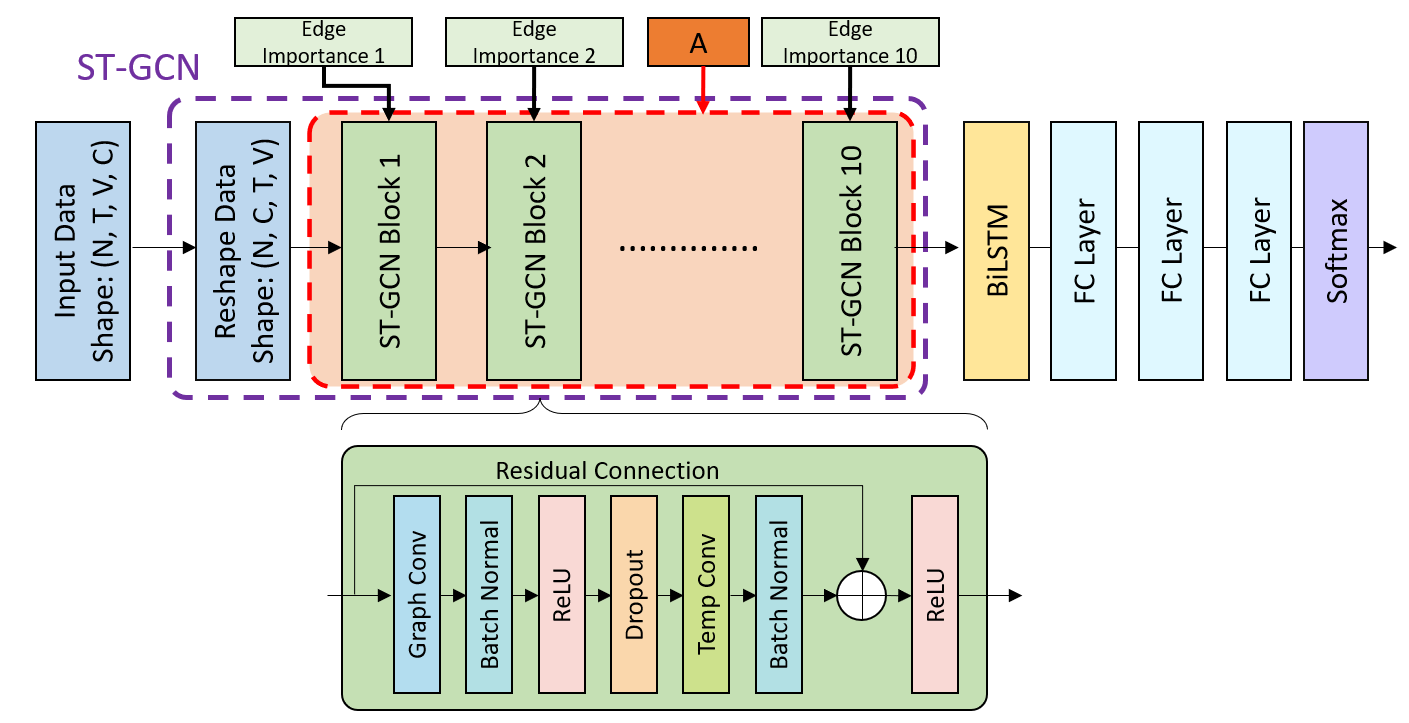}
  \caption{The design of the proposed ST-GCN-BiLSTM model, along with the specific configuration of each individual ST-GCN block.} 
  \label{stgcn model}
\end{figure*}

\subsection{Deep Learning Model}
We introduced a deep learning neural network that combines the ST-GCN model \cite{yan2018spatial} with a BiLSTM framework to achieve the desired task. The ST-GCN model captures both spatial and temporal features, while the BiLSTM refines and enhances the sequential predictions. The architecture of the ST-GCN includes 10 ST-GCN blocks, as illustrated in Fig. \ref{stgcn model}. Each block is equipped with a unique edge importance matrix that is trainable, allowing the model to prioritize keypoints that are most relevant. The adjacency matrix $A$, used in the model, is a fixed matrix derived based on spatial configuration partitioning \cite{yan2018spatial}, and the lower lip node (index 90) from the mouth region was chosen as the root node.

\begin{table}[b]
\caption{Model configuration summary}
\label{table: model summary}
\begin{center}
\scalebox{0.75}{
\begin{tabular}{ccccccc}
\hline
Type  & Kernel number  & Dilation  & Dropout&Output \\ \hline
ST-GCN Block1  & 64  & 1  & 0.15 & (N, 64, T, V)           \\
ST-GCN Block2  & 64  & 1  & 0.15 & (N, 64, T, V)    \\ 
ST-GCN Block3  & 64  & 2  & 0.15 & (N, 64, T, V)    \\
ST-GCN Block4  & 64  & 2  & 0.15 & (N, 64, T, V)    \\ 
ST-GCN Block5  & 128 & 4  & 0.15 & (N, 128, T, V)     \\
ST-GCN Block6  & 128 & 4  & 0.3  & (N, 128, T, V)   \\
ST-GCN Block7  & 128 & 8  & 0.3  & (N, 128, T, V)   \\ 
ST-GCN Block8  & 256 & 8  & 0.3  & (N, 256, T, V)  \\
ST-GCN Block9  & 256 & 16 & 0.3  & (N, 256, T, V)   \\ 
ST-GCN Block10 & 256 & 16 & 0.3  & (N, 256, T, V)    \\
Reshape & -&-  &-  &(N, T, 256 * V)   \\
BiLSTM & 256 & -&-  &(N, T, 512)   \\
Dense & 128 & -&  &(N, T, 128) \\
Dense & 128 &-&-  &(N, T, 128) \\
Dense & 3 & -&-  &(N, T, 3)  \\
\hline
\end{tabular}}
\end{center}
\end{table}

\begin{table*}[t]
\caption {Comparison of segment-wise predictions for eating and drinking gesture recognition using ST-GCN-BiLSTM with dilated and basic ST-GCN block on the OREBA dataset.}
\label{table dilation}
\begin{center}
\scalebox{0.8}{
\begin{threeparttable} 
\begin{tabular}{c|l|ccc|ccc|ccc} 
\toprule
\hline
\multirow{3}{*}{Gesture Type}& \multirow{3}{*}{ST-GCN Type}  & Precision  & Recall & F1-score   & Precision  & Recall  & F1-score  & Precision  & Recall &  F1-score  \\
&&\multicolumn{1}{c}{(\%)}& \multicolumn{1}{c}{(\%)}& \multicolumn{1}{c|}{(\%)}& \multicolumn{1}{c}{(\%)}& \multicolumn{1}{c}{(\%)}& \multicolumn{1}{c|}{(\%)}& \multicolumn{1}{c}{(\%)}& \multicolumn{1}{c}{(\%)}& \multicolumn{1}{c}{(\%)} \\
 &&\multicolumn{3}{c|}{$k = 0.1$}& \multicolumn{3}{c|}{$k = 0.25$}& \multicolumn{3}{c}{$k = 0.5$}\\ 
\midrule
\multirow{2}{*}{Eating}& ST-GCN Basic & 88.18 & 79.04 & 83.36 & 87.01 & 77.99 & 82.26   & 80.78 & 72.41 & 76.37\\ 
& ST-GCN Dilated & 86.70 & 85.66 & 86.18 & 85.59 & 84.47 & 85.03   & 81.84 & 79.01 & 80.40\\
\midrule
\multirow{2}{*}{Drinking}& ST-GCN Basic & 95.91 & 60.26 & 74.02 & 92.87 & 58.44 & 71.74 & 81.63 & 51.28 & 62.99\\ 
& ST-GCN Dilated & 75.32 & 74.36 & 74.84 & 73.33 & 71.43& 72.37   & 66.67 & 64.20 & 65.41\\ 
\hline
\bottomrule
\end{tabular} 
\end{threeparttable} 
}
\end{center}
\end{table*}

\begin{table*}[t]
\caption{Comparison of segment-wise predictions for eating and drinking gesture recognition based on different combinations of skeletal parts using the OREBA dataset.}
\label{table skeleton}
\begin{center}
\scalebox{0.8}{
\begin{threeparttable} 
\begin{tabular}{c|l|ccc|ccc|ccc}
\toprule
\hline
\multirow{3}{*}{Gesture Type}& \multirow{3}{*}{Combination}  & Precision  & Recall & F1-score   & Precision  & Recall  & F1-score  & Precision  & Recall &  F1-score  \\
&&\multicolumn{1}{c}{(\%)}& \multicolumn{1}{c}{(\%)}& \multicolumn{1}{c|}{(\%)}& \multicolumn{1}{c}{(\%)}& \multicolumn{1}{c}{(\%)}& \multicolumn{1}{c|}{(\%)}& \multicolumn{1}{c}{(\%)}& \multicolumn{1}{c}{(\%)}& \multicolumn{1}{c}{(\%)} \\
 &&\multicolumn{3}{c|}{$k = 0.1$}& \multicolumn{3}{c|}{$k = 0.25$}& \multicolumn{3}{c}{$k = 0.5$}\\ 
\midrule
\multirow{6}{*}{Eating}& mouth+hand & 80.52 & 75.09 & 77.71 & 78.65 & 73.34 & 75.90   & 72.16 & 67.29 & 69.64\\ 
& arm+face & 80.52 & 75.09 & 77.71 & 78.65 & 73.34 & 75.90 & 72.16 & 67.29 & 69.64\\ 
& mouth+hand+face & 86.16 & 83.35 & 84.73 & 84.95 & 82.18 & 83.55   & 78.22 & 75.67 & 76.92\\ 
& mouth+hand+arm & 84.54 & 74.51 & 79.21 & 82.56 & 72.75 & 77.35   & 75.03 & 66.12 & 70.30\\  
& mouth+arm+face & \textbf{87.28} & 83.12 & 85.15 & 85.57 & 81.49 & 83.48   & 79.83 & 76.02 & 77.88\\ 
& all four parts & 86.70 & \textbf{85.66} & \textbf{86.18} & \textbf{85.59} & \textbf{84.47} & \textbf{85.03}   & \textbf{81.84} & \textbf{79.01} & \textbf{80.40}\\ 
\midrule
\multirow{6}{*}{Drinking}& mouth+hand & 63.29 & 64.10 & 63.69 & 59.96 & 57.69 & 57.32   & 53.16 & 53.85 & 53.50\\ 
& arm+face & 63.10 & 67.95 & 65.43 & 60.71 & 65.38 & 62.96   & 57.14 & 61.54 & 59.26\\ 
& mouth+hand+face & \textbf{76.36} & 53.85 & 63.16 & \textbf{74.55} & 52.56 & 61.65   & \textbf{67.27} & 47.44 & 55.64\\ 
& mouth+hand+arm & 65.38 & 65.38 & 65.38 & 62.82 & 62.82 & 62.82 & 52.56  & 52.56 & 52.56\\ 
& mouth+arm+face & 74.67 & 71.80 & 73.20 & 73.33 & 70.51 & 71.90   & 61.33 & 58.97 & 60.13\\ 
& all four parts& 75.32 & \textbf{74.36} & \textbf{74.84} & 73.33 & \textbf{71.43} & \textbf{72.37}   & 66.67 & \textbf{64.20} & \textbf{65.41}\\ 
\hline
\bottomrule
\end{tabular} 
\end{threeparttable} 
}
\end{center}
\end{table*}

In each ST-GCN block, graph convolution and temporal convolution are applied to the features sequentially. Following these convolution layers, batch normalization is used to normalize the intermediate feature dimensions. The activation function chosen is ReLU. Additionally, a Dropout layer is applied after the first activation function to mitigate overfitting and enhance the model's generalization capability. A residual connection is also incorporated, linking the input and output of the block to address gradient vanishing issue.

The original ST-GCN model proposed by \cite{yan2018spatial} includes a temporal convolution step. However, it uses a fixed kernel size for the temporal convolution, essentially performing a 1D-CNN along the time dimension with a static kernel size. To expand the receptive field, this study employs a non-causal TCN model with dilated convolutions. To differentiate between the original TCN model and the version using dilation, the original ST-GCN with fixed temporal kernel size is referred to as ``ST-GCN Basic," while the version incorporating dilated temporal convolutions is termed ``ST-GCN Dilated" in this research.

The input format of the deep learning model, based on the skeleton partition strategy and sliding window approach, is represented as $(N, T, V, C)$. Here, $N$ denotes the batch size, $T$ represents the time series length (equal to the sliding window length), $V$ indicates the number of skeleton keypoints per person (23), and $C$ corresponds to the number of input channels. In this study, $C$ is set to 3, incorporating the X and Y coordinate positions along with the confidence score. Table \ref{table: model summary} provides an overview of the model architecture.

Based on experiments, the window length is set to 6 s, represent 144 frames per window for OREBA dataset. The model was trained for 50 epochs with a batch size of 64. An Adam optimizer was employed, configured with a learning rate of 0.0005. The loss function used in the model combines classification loss and smoothing loss. The classification component is based on cross-entropy loss, while the smoothing loss is calculated using a truncated mean squared error (MSE). Further details regarding the combined loss function can be found in \cite{b37}. All experiments were conducted using an NVIDIA A100 GPU provided by the Vlaams Supercomputer Centrum (VSC).

\subsection{Evaluation}
The output of the model is frame-wise prediction, however, frame-wise evaluation method cannot reflect the the detection performance on gesture level. To this end, the segment-wise evaluation method \cite{10596709} has been used. The segment-wise evaluation method first calculates the Intersection Over Union (IoU) between ground truth intake gesture and predicted gesture along temporal dimension. Then a defined IoU threshold $k$ is used to define the segment-wise TP, FP, and FN. Three IoU thresholds ($k=0.1,0.25,0.5$) were selected in this study. Such a method also punishes oversegmentation and merge error. Subsequently, the segment-wise F1-score can be calculated. Detailed information of such evaluation scheme can be found in \cite{10596709}. 

\begin{figure*}[t]
  \centering
  \includegraphics[scale=0.5]{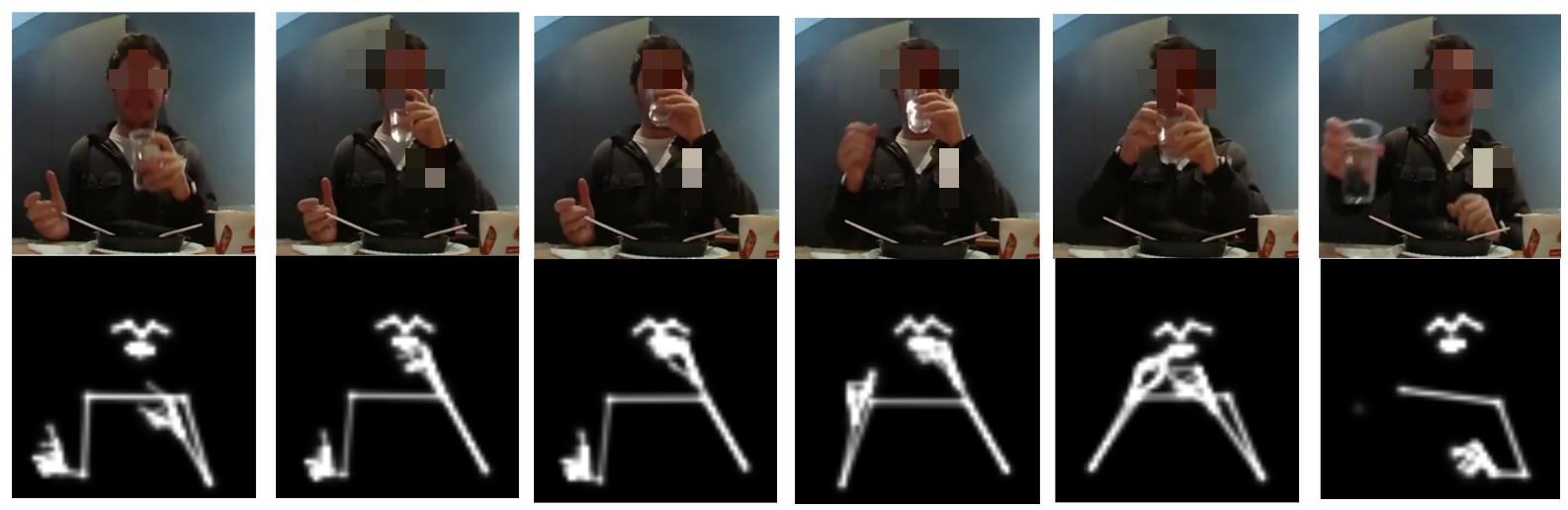}
  \caption{An example of a drinking gesture is shown, where the model incorrectly classifies it as an eating gesture. This misclassification might be due to the change in hand position while placing the glass down.} 
  \label{drinking figure}
\end{figure*}

\begin{figure*}[t]
  \centering
  \includegraphics[scale=0.5]{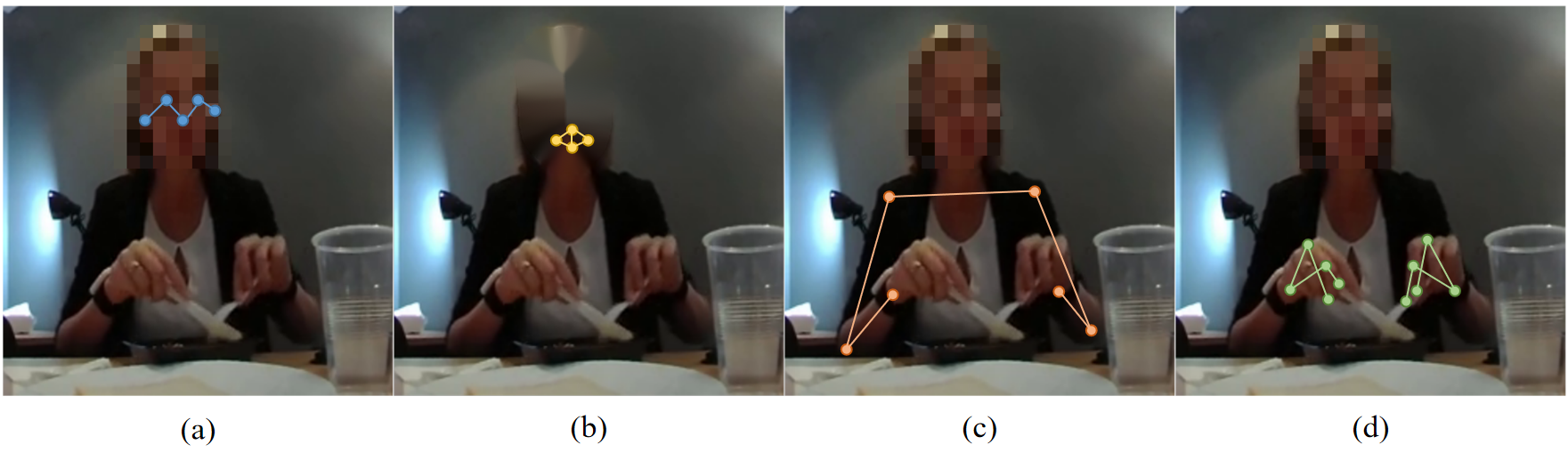}
  \caption{The figure illustrates the skeletal representations of four distinct body parts: (a) face, (b) mouth, (c) arm, and (d) hands. Each subfigure highlights the skeletal structure specific to the corresponding body part. When multiple body parts are utilized by the model, suitable connections between them are incorporated.} 
  \label{body part result}
\end{figure*}

\begin{table}[t]
\caption{Segment-wise prediction performance of eating and drinking gesture recognition on Smartphone Footage Dataset}
\label{table footage result}
\begin{center}
\scalebox{1}{
\begin{threeparttable} 
\begin{tabular}{c|l|ccc|c}
\toprule
\hline
Gesture type& $k$ &TP(\#)&FP(\#)&FN(\#)&F1-score (\%)\\
\midrule
\multirow{3}{*}{Eating}
& 0.1  & 705 & 109 & 132 & 85.40\\ 
& 0.25  & 693 & 114 & 139 & 84.56\\ 
& 0.5  & 625 & 149 & 172 & 79.57\\
\midrule
\multirow{3}{*}{Drinking}
& 0.1  & 40 & 10 & 28 & 67.80\\ 
& 0.25  & 38 & 11 & 29 & 65.52\\ 
& 0.5  & 35 & 11 & 32 & 61.95\\
\hline
\bottomrule
\end{tabular} 
\end{threeparttable} 
}
\end{center}
\end{table}

\section{Results and Discussion}
The OREBA dataset was utilized to evaluate the effectiveness of the proposed method. Both the ST-GCN-BiLSTM Basic and ST-GCN-BiLSTM Dilated models were assessed, as summarized in Table \ref{table dilation}. The ST-GCN-BiLSTM Dilated model outperformed the Basic model, achieving F1-scores of 86.18\% and 80.40\% for detecting eating gestures at thresholds $k$ = 0.1 and 0.5, respectively. For drinking gestures, the proposed method attained F1-scores of 74.84\% and 65.41\%. The model also showed lower sensitivity to drinking gestures. This limitation can be attributed to the smaller number of drinking samples and the inconsistent duration of such gestures. Additionally, the distinction between drinking and eating gestures is sometimes ambiguous. An example of a misclassified drinking gesture is illustrated in Fig. \ref{drinking figure}. The model mistakenly identified the gesture as eating, possibly due to the subject’s use of their right hand to place the glass down, resulting in the latter part of the gesture being labeled incorrectly. Incorporating RGB data and employing a multimodal deep learning framework could potentially improve the model's performance in distinguishing between these gestures.

Graph convolution is rooted in graph theory, meaning that the choice of the underlying skeletal representation plays a crucial role in feature extraction. In the OREBA dataset, videos were recorded with a camera positioned on a table, capturing primarily the upper body of participants. In this study, four key body parts (face, mouth, arm, and hands) were selected and combined for analysis, as shown in Fig. \ref{body part result}. However, it remains uncertain whether this combination is the most efficient or if it introduces redundancy. To address this, six different combinations of body parts were tested as skeletal representations for deep learning predictions. As shown in Table \ref{table skeleton}, the model achieved its best performance when all four body parts were included. The combinations of ``mouth + arm + face" and ``mouth + hand + face" produced similar results, with F1-scores of 84.73\% and 85.15\% for eating gesture prediction at an IoU threshold of 0.1. Notably, the ``mouth + hand + face" combination demonstrated a considerable improvement in detecting drinking gestures compared to the other two combinations. The remaining three skeletal combinations performed poorly in predicting both eating and drinking gestures. 

After validating the model's performance on the OREBA dataset, we extended our evaluation to a more complex scenario using our smartphone footage dataset. The model, trained on the OREBA dataset, was tested on this home environment dataset. The results, summarized in Table \ref{table footage result}, show that the model achieved an eating F1-score of 85.40\% at $k=0.1$, and 79.57\% at $k=0.5$. However, consistent with previous findings, the model demonstrated limited sensitivity to drinking gestures. The number of FNs for drinking gestures revealed that a significant number of these actions were not detected by the deep learning model. These findings highlight that while the ST-GCN-BiLSTM Dilated model performs well in identifying eating gestures in diverse and realistic settings, improvements are needed to enhance its ability to accurately detect drinking gestures. It is also noteworthy that our previous study \cite{10034636} used RGB input for deep learning models, with the SlowFast model \cite {feichtenhofer2019slowfast} achieving only 59.3\%. Utilizing skeleton-based input in conjunction with deep learning models offers significant advantages for recognizing fine-grained eating-related activities across dynamic and diverse environmental conditions. However, the effectiveness of this approach is highly contingent on the quality of keypoints extracted from raw video data. Consequently, the performance of the skeleton extraction framework serves as a critical bottleneck, directly influencing the overall accuracy and reliability of the proposed method.

\section{Conclusion}
This study introduced a skeleton-based deep learning approach for detecting food intake gestures that is resilient to changes in background, requires minimal data, and ensures lower privacy concern. The proposed pipeline primarily involves extracting skeletal data from video frames and using this data as input for the ST-GCN-BiLSTM Dilated model. Skeleton features were extracted using MMPose, focusing on 23 upper body keypoints, including those from the face, mouth, arms, and hands. The model incorporates a dilated ST-GCN to capture temporal and spatial features from the skeleton data, complemented by a BiLSTM layer to refine prediction outcomes. The ST-GCN-BiLSTM model shows comparable performance compared to video-based RGB signal models on OREBA dataset, achieving better results on the Smartphone dataset under cross-dataset validation. This underscores the effectiveness and robustness of skeleton-based deep learning models for dietary monitoring tasks. Future work will focus on exploring multimodal approaches to further enhance the model’s detection performance.

\section*{Acknowledgment}
The computational resources and services used in this work were provided by the VSC (Flemish Supercomputer Center), funded by the Research Foundation Flanders (FWO) and the Flemish Government – department EWI. 
\bibliographystyle{IEEEtran}
\bibliography{papers}
\end{document}